# Enhancing Large Language Models with Domain-specific Retrieval Augment Generation:

## A Case Study on Long-form Consumer Health Question Answering in Ophthalmology

Aidan Gilson, BS[1,*], Xuguang Ai, MS[2,*], Thilaka Arunachalam, BS[3], Ziyou Chen, MD[4], Ki Xiong Cheong, MD[4], Amisha Dave, MD[1], Cameron Duic, MD[3], Mercy Kibe, MD[1], Annette Kaminaka, MD[3], Minali Prasad, MD[3], Fares Siddig, MD[3], Maxwell Singer, MD, PhD[1], Wendy Wong, MD[4], Qiao Jin, MD, PhD[5], Tiarnan D.L. Keenan, BM BCh, PhD[3], Xia Hu, PhD[6], Emily Y. Chew, MD[3], Zhiyong Lu, PhD[5], Hua Xu, PhD[2], Ron A. Adelman, MD, MPH[1], Yih-Chung Tham, PhD[7,+], Qingyu Chen, PhD[2,+]

1. Department of Ophthalmology and Visual Science, Yale School of Medicine, Yale University, New Haven, USA

2. Department of Biomedical Informatics and Data Science, Yale School of Medicine, Yale University, New Haven, USA

3. Division of Epidemiology and Clinical Applications, National Eye Institute, National Institutes of Health, Bethesda, USA

4. Singapore Eye Research Institute, Singapore National Eye Centre, Singapore

5. National Center for Biotechnology Information, National Library of Medicine, National Institutes of Health, Maryland, USA

6. Department of Computer Science, Rice University, Houston

7. Department of Ophthalmology, Yong Loo Lin School of Medicine, National University of Singapore, Singapore

*These authors contributed equally to this work.

+Corresponding authors: thamyc@nus.edu.sg, qingyu.chen@yale.edu.

# Abstract

Despite the potential of Large Language Models (LLMs) in medicine, they may generate responses lacking supporting evidence or based on hallucinated evidence. While Retrieval Augment Generation (RAG) is popular to address this issue, few studies implemented and evaluated RAG in downstream domain-specific applications.

We developed a RAG pipeline with ~70,000 ophthalmology-specific documents that retrieve relevant documents to augment LLMs during inference time. In a case study on long-form consumer health questions, we systematically evaluated the responses – including over 500 references – of LLMs with and without RAG on 100 questions with 10 healthcare professionals. The evaluation focuses on factuality of evidence, selection and ranking of evidence, attribution of evidence, and answer accuracy and completeness.

LLMs without RAG provided 252 references in total. Of which, 45.3% hallucinated, 34.1% consisted of minor errors, and 20.6% were correct. In contrast, LLMs with RAG significantly improved accuracy (54.5% being correct) and reduced error rates (18.8% with minor hallucinations and 26.7% with errors). 62.5% of the top 10 documents retrieved by RAG were selected as the top references in the LLM response, with an average ranking of 4.9. The use of RAG also improved evidence attribution (increasing from 1.85 to 2.49 on a 5-point scale, P<0.001), albeit with slight decreases in accuracy (from 3.52 to 3.23, P=0.03) and completeness (from 3.47 to 3.27, P=0.17).

The results demonstrate that LLMs frequently exhibited hallucinated and erroneous evidence in the responses, raising concerns for downstream applications in the medical domain. RAG substantially reduced the proportion of such evidence but encountered challenges. In contrast to existing studies, the results highlight that (1) LLMs may not select top-ranked documents by RAG, which results in hallucinated evidence remaining, (2) LLMs may miss top-ranked documents by RAG, and (3) irrelevant documents by RAG downgrade response accuracy and completeness, especially in challenging tasks such as long-form question answering.

In conclusion, in long-form medical question answering, the RAG approach demonstrated improved effectiveness over non-RAG approach. Nevertheless, there are still challenges in evidence retrieval, selection, and attribution, highlighting the need for further development in domain-specific LLM and RAG techniques.

## Introduction

Large Language Models (LLMs) represent one of the latest advancements in AI systems designed for language modeling[1,2]. Compared with early language models, LLMs demonstrate notable capabilities in natural language generation and reasoning tasks such as reading comprehension[3], translation[4], and question-answering[5]. Studies also demonstrate that LLMs exhibit in-context learning abilities, enabling them to effectively interpret and generate text even when provided with minimal prompts (zero-shot learning) or a limited number of example demonstrations (few-shot learning)[6]. In the medical domain, LLMs also demonstrate potential across a range of applications[7-12]. We conducted a systematic evaluation of the effectiveness of LLMs across 12 biomedical natural language processing benchmarks demonstrating that LLMs already surpassed previous state-of-the-art methods in generative applications under zero/few-shot scenarios. Additionally, other studies have shown promising performance of LLMs in disease diagnosis[13], impression generation[14], and medication education[15].

Despite this promise, studies also highlight the issue of hallucination in medical applications of LLMs, where these models may produce responses that are linguistically fluent and semantically coherent but may deviate from factual accuracy or contain fabricated information[16-18]. For instance, Hou *et al.* manually examined over 10,000 LLM-generated responses to 600 biomedical and genomic questions across six topics[19]. They found that responses were often entirely hallucinated. Similar issues have been reported in biomedical information retrieval[20] and biomedical relation extraction studies[21]. In response, Retrieval Augmented Generation (RAG) has been proposed to address such issue[16,22-24]. The basic concept of RAG involves retrieving top the most relevant documents based on user queries and using these documents to augment LLMs for generating responses. A typical RAG pipeline includes indexing targeted documents, employing retrieval functions such as standard BM25 or those based on semantic similarity to identify relevant documents, and augmenting the top retrieved documents to LLMs. It offers two primary advantages: (1) it does not require pretraining and (2) it can be updated directly with the latest domain-specific knowledge.

Reviews and perspectives on LLMs in biomedicine and healthcare emphasize the importance of RAG in addressing hallucinations and providing the latest domain-specific knowledge to LLMs without the need for retraining[25,26]. However, to date, only a few studies have implemented RAG in specific downstream applications. For example, Guo et al. utilized RAG for biomedical lay summary generation[17]. They curated PubMed abstracts and their corresponding author-drafted lay language summaries from 12 journals to augment LLMs for lay summaries and simplification[17]. Additionally, we conducted a pilot study on augmenting LLMs with domain-specific tools to answer biological questions[27]. Other medical applications employing RAG include medical question answering and text summarization[28-30]. Nevertheless, three limitations persist. First, while RAG could effectively address hallucinations, it may also introduce trade-offs with the quality of generated responses, as it may retrieve irrelevant documents and consequently downgrade LLM responses. Second, in biomedical and health domains, evidence attribution, i.e., providing supporting evidence that can be verified and traced back to a statement, is arguably more critical than merely providing an accurate response. For instance, in medical question answering, answers not only need to be correct but also require accurate references for healthcare professionals as justification. To date, limited evaluations have been conducted on the factuality and relevance of evidence. Last, there is a scarcity of domain-specific RAG applications for LLMs. For example, despite the importance and potential use cases of LLMs in ophthalmology, we are aware of only one existing study that applied RAG to improve the accuracy of multiple-choice questions[31].

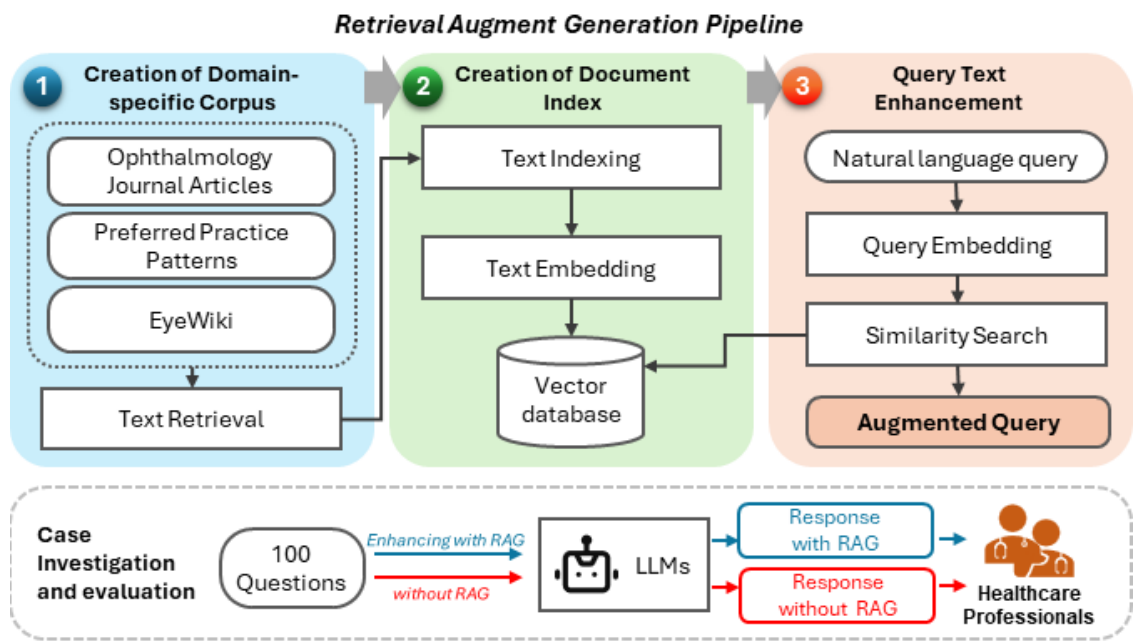

Figure 1. Study overview. A Retrieval Augment Generation (RAG) with ~70K domain-specific corpora was developed. A case study evaluates the LLM responses with and without RAG on 100 consumer health questions on factuality of evidence, selection of evidence, response accuracy, completeness, and evidence attribution reviewed by healthcare professionals.

In response to these challenges, we developed an ophthalmology-specific Retrieval Augment Generation (RAG) approach and conducted a systematic evaluation on the case of long-form consumer health question answering, as shown in Figure 1. The primary contributions of this work are three-fold:

First, we curated approximately 70,000 domain-specific documents, including biomedical literature, clinical practice guidelines, and relevant wiki articles, and implemented a RAG pipeline. This pipeline recommends the most relevant documents for LLMs to augment the responses.

Second, we systematically evaluated over 500 references for 100 consumer health questions answered by LLMs with and without RAG and examined the factuality of the references and the selection of most relevant documents by RAG. 10 healthcare professionals (six medical students, three residents, and one attending specialist) double-blindly evaluated the accuracy and completeness of the responses and evidence attribution for 30 questions with and without RAG, resulting in 540 annotations per annotator.

Third, we also made the related data, models, and codes available to the community via https://github.com/qingyu-qc/medical_rag_evidence for reproducibility and further development.

Table 1. Domain-specific corpora for retrieval augment generation

| Domain-specific data | Size |
| --- | --- |
| PubMed Abstracts from Ophthalmology Journals | 66,269 abstracts |
| Preferred Practice Patterns | 24 documents |
| Eyewiki | 1,494 articles |

## Data and Methods

Figure 1 demonstrates the overview of the study. The details are below.

### Retrieval Augment Generation Pipeline

**Creation of Domain-Specific Corpus**. Initially, we curated a corpus of approximately 70,000 ophthalmology-specific documents from three primary resources, as outlined in Table 1.

**Ophthalmology Journal Articles**. We sourced articles from ten primary ophthalmology journals, including *Ophthalmology, JAMA Ophthalmology, American Journal of Ophthalmology, British Journal of Ophthalmology, Retina, Ophthalmology Glaucoma, Journal of Cataract and Refractive Surgery, Asia-Pacific Journal of Ophthalmology, Investigative Ophthalmology and Visual Science, and Survey of Ophthalmology*, published since 1990. Abstracts and related metadata such as journal names, publication years, and DOIs were extracted using e-utils. Further, we conducted quality control and removed abstracts with invalid metadata. Each abstract was treated as a single document in the corpus.

**Preferred Practice Patterns**. We further collected Preferred Practice Patterns in Ophthalmology from the American Academy of Ophthalmology (AAO): These guidelines are publicly accessible via the AAO website and offer expert panel-developed recommendations for high quality eye care. The guidelines are updated every five years and may span hundreds of pages. Each page was processed as a single document in the corpus.

**EyeWiki**. This collection comprises publicly accessible articles written by ophthalmologists. Aimed at providing introductory and educational materials using simpler language, the articles cover various topics on eye diseases, diagnoses, and treatments. Each page was treated as a single document in the corpus.

**Indexing, Embedding, and Querying.** The documents underwent further segmentation into 1024-token snippets for indexing. Text snippet embeddings (semantic representations) were generated using text-embedding-ada-002[32] and stored in the database. Given a natural language query, the RAG pipeline generates its embedding, performs dense retrieval, identifies the top similar candidates based on cosine similarity of embeddings, and augments those candidates to an LLM to generate a response.

## Case Investigation and Evaluations

**Task**. We evaluated the effectiveness of RAG in augmenting LLMs for long-form question answering in ophthalmology. In contrast to binary or multiple-choice questions, long-form question answering involves providing a free-text passage with reasoning and supporting evidence to justify the answer. We chose 100 publicly accessible question-answer pairs from the Ask An Ophthalmologist forum by AAO, by sampling 20 each from the following five topics: Retina, Glaucoma, Cataract, Dry Eye, and Uveitis. These questions cover various aspects of eye health, vision problems, ophthalmic conditions, and eye care.

Each question prompted the LLM as follows: "Answer this question and provide references at the end of your response. The references should adhere to the AMA format," Same prompts were used for the LLMs with and without RAG. When using RAG, it retrieves the top 10 most relevant documents in the corpus to augment the LLM. We used GPT-3.5 (gpt-3.5-turbo-0613) as the representative LLM. We chose GPT-3.5 as the representative LLM for its reasonable accuracy and cost efficiency[10]. Note that RAG can be integrated into any LLMs. Each LLM with and without RAG will provide different references and answers, which adds substantially to the manual evaluation task. The focus of the paper is to assess the evidence factuality, evidence selection, and evidence attribution, not the performance of LLM itself. The temperature of the LLM was set to 0 to minimize the variance of generated responses. We also manually verified that providing top 10 most relevant documents did not reach the token limit of GPT-3.5.

**Evaluation of Factuality of Evidence**. We manually examined the top three references in the LLM responses, categorizing them as (1) correct references, where the references are real with correct metadata, (2) references with minor errors, where the references are real but have minor metadata errors, or (3) hallucinated references, where the references do not exist.

**Evaluation of the Selection and Ranking of Retrieved Documents**. For the top 10 most relevant documents retrieved by RAG, we quantified how many were selected as the top three references in the LLM responses and the average rankings of the selected documents. For instance, an LLM response has three references. Two of them are in the top 10 relevant documents retrieved by RAG, with rankings of 3 and 5. In this case, 66% of the top-ranked documents by RAG were selected as top references in the LLM response, with an average ranking of 4.

**Evaluation of Response Accuracy, Completeness, and Evidence Attribution**. We further sub-sampled 30 questions, 10 each for retina, glaucoma, and cataract, representing three major ophthalmology subspecialties. Ten healthcare professionals (six medical students, three residents, and one attending specialist) manually evaluated the responses, with and without RAG, in a blinded manner with shuffled orders. Each response was rated on three axes: accuracy, completeness, and evidence attribution, on a scale from 1 (poor) to 5 (perfect). The details of the healthcare professionals and evaluation guidelines are provided in Supplementary Materials.

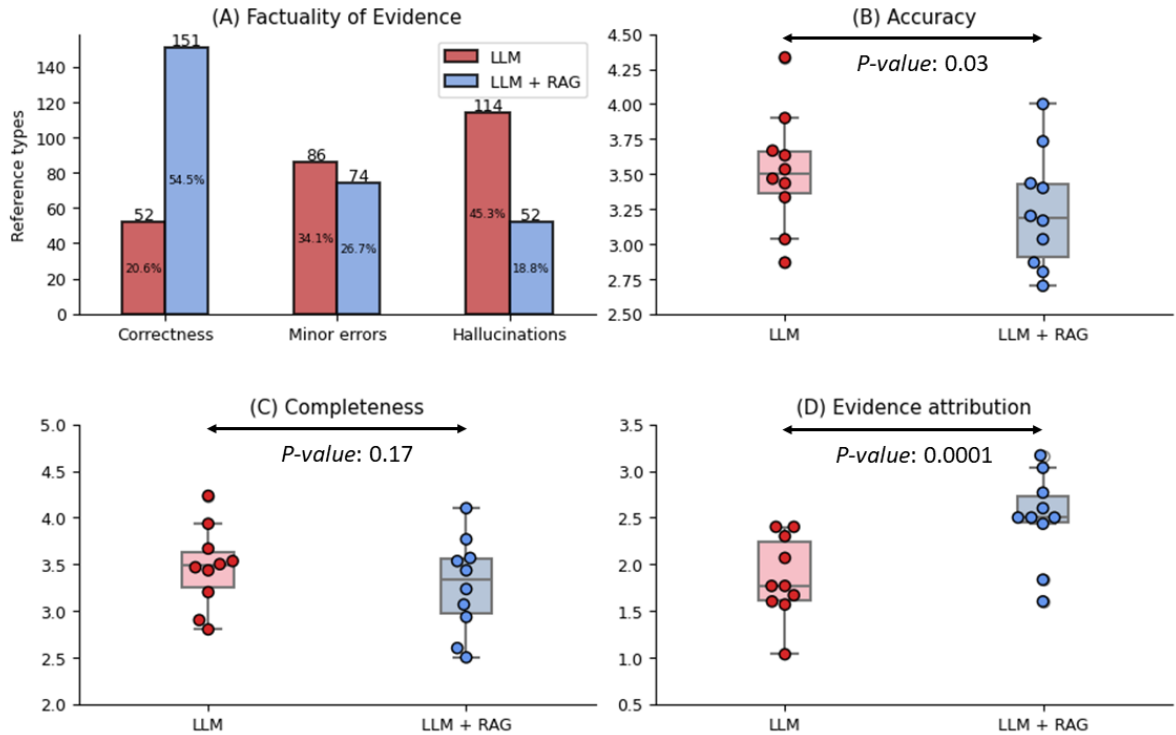

Figure 2. Evaluation results on LLMs with and without RAG. (A) Factuality of references: correct references, where the references are real with correct metadata, references with minor errors, where references are real but have minor metadata errors, and hallucinated references, where the references do not exist. (B)-(D) Average manual evaluation scores on response accuracy, response completeness, and evidence attribution, respectively.

## Results

### Evaluation of Factuality of Evidence

Figure 2(A) shows the evaluation results. For the 100 questions, the LLM without RAG and with RAG provided 252 and 277 references in the responses, in total, respectively. Of the 252 references of the LLM without RAG, 20.6% (52) were correct references (i.e., real references with correct metadata), 34.1% (86) were references with minor errors (i.e., real references but with minor metadata errors), and 45.3% (114) were hallucinated references.

In contrast, out of the 277 references in the responses the LLM with RAG, 54.5% (151) were correct references, 26.7% (74) were references with minor errors, and 18.8% (52) were hallucinated references. The proportion of correct references also increased from 20.6% to 54.5%. This shows that RAG can dramatically improve the factuality of evidence of LLMs. However, the final responses of the LLM still had hallucinated references, at 18.8%, which implies that the LLM may not include the top retrieved documents by RAG in the final responses. We describe the results below.

## Evaluation of the Selection and Ranking of Retrieved Documents

In total, of the 277 references provided in the LLM + RAG responses, 173 references were from the top 10 most relevant documents retrieved by RAG. In other words, only 62.5% of the references identified by RAG were selected as top references in the final LLM responses. In addition, the average ranking of those top references was 4.89 (standard deviation 2.40), with median ranking of 4.67. This implies that the top-ranked documents by RAG may not be selected by LLM in the final response; the selected documents in the final responses are at the median rankings by RAG.

Table 2. Quantitative comparisons on response accuracy, response completeness, and evidence attribution. Both overall results and results per specific topic are summarized.

| | LLMs without RAG | LLMs with RAG | P-value |
|---|---|---|---|
| **Overall questions** | | | |
| Accuracy | 3.52 | 3.23 | 0.035 |
| Completeness | 3.47 | 3.27 | 0.178 |
| Evidence attribution | 1.86 | 2.48 | 0.000 |
| **Cataract questions** | | | |
| Accuracy | 3.91 | 3.41 | 0.006 |
| Completeness | 3.77 | 3.64 | 0.419 |
| Evidence attribution | 1.94 | 2.71 | 0.000 |
| **Glaucoma questions** | | | |
| Accuracy | 3.46 | 3.39 | 0.695 |
| Completeness | 3.39 | 3.18 | 0.587 |
| Evidence attribution | 1.76 | 2.76 | 0.000 |

**Retina questions**

| | | | |
|---|---|---|---|
| Accuracy | 3.21 | 3.35 | 0.092 |
| Completeness | 2.90 | 3.05 | 0.069 |
| Evidence attribution | 1.87 | 2.02 | 0.267 |

**(A) Can I have LASIK if I am a glaucoma suspect?**

| LLM | It is generally not recommended for individuals who are glaucoma suspects to undergo LASIK surgery [...] | Wrong answers |
| | | |
| | A study published in the Journal of Glaucoma found that LASIK can cause a significant increase in IOP in glaucoma suspects, which can persist for up to 6 months after surgery. The study concluded that LASIK should be avoided in individuals with glaucoma or those who are at risk of developing glaucoma. | Evidence based on hallucinated references |
| | | |
| | Another study published [...] | |
| | | |
| | References: | Hallucinated references |
| | 1. Lee JW, Yau GS, Yuen CY, Wong RL, Yuen HK. Intraocular pressure changes after laser [...] | |
| | 2. Lee JH, Kim EK, Kim HB, Kim TI. Intraocular pressure changes after laser in [...] | |

| LLM + RAG | Yes, you may be able to have LASIK if you are a glaucoma suspect. However, it is important to discuss the risks and benefits of the procedure [...] | Correct answers |
| | | |
| | Glaucoma patients present a unique set of challenges to physicians performing corneal refractive surgery [...] Patients undergo a transient but significant rise in intraocular pressure during the laser-assisted in situ keratomileusis (LASIK) procedure with risk of further optic nerve damage or retinal vein occlusion. Glaucoma patients with filtering blebs are also at risk of damage to the bleb by the suction ring. Steroids, typically used after refractive surgery, can increase intraocular pressure in steroid responders, which is more prevalent among glaucoma patients [...] | Evidence based on actual references |
| | | |
| | References: | References exist and relevant |
| | 1. Considerations of glaucoma in patients undergoing corneal refractive surgery [...] | |
| | 2. Primary Open-Angle Glaucoma Preferred Practice Pattern [...] | |

**(B) After many injections for wet AMD, my husband stopped treatment during the pandemic and has lost central vision. Would further injections bring back his central vision or stop further loss?**

| LLM | It is possible that further injections for wet AMD could bring back central vision or stop further loss, but this would depend on the individual case [...] | Correct answers |
| | | |
| | Studies have shown that regular injections of anti-VEGF medications can improve visual acuity and reduce the risk of vision loss in patients with wet AMD (1, 2) [...] | Not supported by the references |
| | | |
| | References: | Somewhat relevant references |
| | 1. Brown DM, Michels M, Kaiser PK, et al. Ranibizumab versus verteporfin [...] | |
| | 2. Rosenfeld PJ, Brown DM, Heier JS, et al. Ranibizumab for neovascular [...] | Hallucinated references |
| | 3. Boyer DS, Schmidt-Erfurth U, van Lookeren Campagne M, et al. Effect of cessation and restarting of anti-vascular [...] | |

| LLM + RAG | It is possible that further injections may help to bring back your husband's central vision or stop further loss. However, it is important to note that the efficacy of anti-VEGF injections in restoring vision is highly variable [...] | Correct answers |
| | | |
| | Studies have shown that patients with neovascular age-related macular degeneration (AMD) who experience acute vision loss despite periodic injections of anti-VEGF treatment may recover their vision to pre-loss levels at least once, but the final best-corrected visual acuity (BCVA) is often worse than that of patients who did not experience acute vision loss.1,2 [...] | Evidence based on actual references |
| | | |
| | References: | References exist and relevant |
| | 1. Lai TY, Chen YJ, Chen YH, et al. Investigation of incidence and causes of acute vision loss during anti-vascular endothelial [...] | |
| | 2. Schmitz-Valckenberg S, Holz FG, Bird AC, et al. Five-year visual acuity outcomes and injection patterns in patients with pro re nata treatments [...] | |
| | 3. Chen YJ, Lai TY, Chen YH, et al. Impact of COVID-19 pandemic lockdowns on visual acuity of patients with neovascular AMD: a large cohort. [...] | |

**(C)** What are the signs and symptoms of eye microaneurysms?

LLM + RAG: Signs and symptoms of eye microaneurysms can vary depending on the underlying cause. Generally, patients may experience a decrease in central vision, decreased color vision, or may be asymptomatic. [...] Additionally, fluorescein angiography may reveal disc edema and leakage and [...] Additionally, macroaneurysms can occur secondary to a branch retinal vein occlusion, diabetic retinopathy, radiation retinopathy [...]

In cases of premacular hemorrhage, YAG laser can be used to perform a posterior hyaloidotomy and release the hemorrhage into the vitreous for faster clearance. Recent studies have also shown promising results using anti-VEGF agents such as bevacizumab and ranibizumab in patients with macroaneurysm-associated macular edema.

References:
1. Boyd K, Janigian RH. Macular Pucker. American Academy of Ophthalmology [...]
2. Optical coherence tomographic characteristics of microaneurysms [...]
3. Porter D, Vemulakonda GA. Blood Pressure [...]

*Reference 2 is not discussed in the response*

*Irrelevant reference*
*Somewhat relevant reference*
*Irrelevant reference*

Figure 3. Case analysis. The content in square brackets is omitted due to space limitation.

## Evaluation of Response Accuracy, Completeness, and Evidence Attribution

Figure 2(B)-(D) shows the evaluation results by 10 healthcare professionals. The RAG significantly improved evidence attribution, from a mean 1-5 ranking of 1.85 to 2.49 (P-value 0.0001). However, it also had a trade-off, with small decreases in accuracy (from 3.52 to 3.23, P-value 0.03) and completeness (from 3.47 to 3.27, P-value 0.17). Table 2 further shows the detailed subtopic results for cataract, glaucoma, and retina. These results are consistent with the overall results. We have three key observations. First, the LLM without RAG may provide responses with reasonable accuracy (e.g., mean of 3.46 for the glaucoma questions). However, it may generate hallucinated and irrelevant references that contains misleading information. This makes it challenging for downstream users to justify the correctness of the responses. Second, although RAG consistently improved the evidence attribution, the evidence attribution score remains significantly lower than the accuracy and completeness scores. This underscores the need for better approaches to further improve the relevance of the evidence without compromising accuracy and completeness. Third, domain-specific long-form question answering is still challenging for LLMs. Compared with existing studies reporting over 80%

accuracy in multiple choice question answering, both accuracy and completeness scores were substantially lower.

We further manually analyzed the cases and presented three representative examples in Figure 3. In Case (A), the LLM without RAG produces incorrect answers supported by non-existent studies, whereas the LLM with RAG delivers accurate answers substantiated by relevant references. In Case (B), both LLMs with and without RAG yield accurate answers, yet the LLM without RAG bases its claims on non-existent studies and some irrelevant references. In Case (C), illustrating the challenges of the LLM with RAG, it provides three references, two of which are irrelevant. The responses fail to synthesize the relevant references.

## Discussions

### Main Findings and Interpretations

In this study, we developed a domain-specific RAG pipeline consisting of about 70,000 documents, including biomedical literature, clinical practice guidelines, and relevant wiki articles in ophthalmology, and performed a systematic evaluation on the case study of long-form consumer health question answering with 10 healthcare professionals. This study contributes three main findings.

First, it demonstrates that LLMs frequently include hallucinated and erroneous evidence while they may generate reasonable answers. The results quantify that almost half of the references are hallucinated and about 30% of references contain errors in the responses of LLMs. This pressing issue needs to be addressed as it concerns the downstream accountability of LLMs in the medical domain.

Second, through a systematic evaluation on LLMs with RAG, the results highlight that RAG could improve factuality and evidence attributions; however, there are three primary challenges to address: (1) LLMs may not always select the documents provided by RAG, resulting in the persistence of hallucinated evidence; (2) LLMs may also miss top-ranked documents identified by RAG; and (3) irrelevant documents identified by RAG can negatively impact response accuracy and completeness.

Third, it also leads to the successful implementation of a domain-specific RAG pipeline, and we make the related data, models, and code available to the community for reproducibility and further development.

## Comparison with Literature

As mentioned, most studies focused on assessing the content level of LLM-generated responses, such as content accuracy, rather than on the evidence level[16]. In the medical domain, however, evidence is arguably more crucial[33]; biomedical researchers and healthcare professionals need to verify evidence and justify claims rather than focusing solely on responses. Pioneering studies found that up to 90% of LLM responses are not supported by the sources they provide[34]. Our case evaluation on real consumer health questions systematically quantifies the factuality of evidence, selection and ranking of evidence, and evidence attribution, in addition to response accuracy and completeness. Additionally, only a few studies have implemented RAG in specific downstream applications. We are aware of only one existing study that applied RAG to improve the accuracy of multiple-choice questions in LLMs within ophthalmology[31]. We implemented by far the largest ophthalmology-specific RAG pipeline. Furthermore, studies in the medical domain have reported higher accuracy when using RAG, but this is often based on simpler tasks such as multiple-choice question answering[23,27,31]. Our results reveal important challenges in long-form question answering for real consumer health questions, particularly where there are no candidate answer options and LLMs may need to synthesize information from multiple documents. The evaluation highlights the three challenges of using RAG to augment LLM responses as detailed above.

## Limitations

Our study also has several potential limitations. First, as manual annotations (540 annotations per participant) are costly, we evaluated only GPT-3.5 as the representative LLM and default RAG configurations as the representative RAG. Each representation will require re-annotation. However, this is arguably the most common choice for downstream users. In the future, we will also evaluate other representative LLMs including GPT-4, LLaMA[35], and PMC-LLaMA[36], as well as explore different RAG setups, such as using domain-specific embeddings for semantic search[37]. Second, we hypothesized that Ophthalmology domain-specific resources are more effective and curated about 70K Ophthalmology domain-specific resources ranging from biomedical literature, clinical guidelines, and educational materials for RAG. However, it might be possible that general domain or medical domain resources also contain relevant documents. We will evaluate the trade-offs for selection of corpora (e.g., using the entire PubMed vs Ophthalmology-specific literature) in the future.

## Acknowledgement

This study was supported by 1K99LM014024 and NIH Intramural Research Program (IRP), National Eye Institute and National Library of Medicine.